%% file: main.tex
\documentclass[10pt,twocolumn,letterpaper]{article}

\usepackage[pagenumbers]{cvpr} 
\input{preamble}

\definecolor{cvprblue}{rgb}{0.21,0.49,0.74}
\usepackage[pagebackref,breaklinks,colorlinks,citecolor=cvprblue]{hyperref}

\def\paperID{*****} 
\def\confName{CVPR}
\def\confYear{2024}

\title{Harnessing Temporal Causality for Advanced Temporal Action Detection}

\author{
Shuming Liu$^{1}$ \quad
Lin Sui$^{2}$ \quad
Chen-Lin Zhang$^{3}$ \quad 
Fangzhou Mu$^{4}$ \quad
Chen Zhao$^{1}$ \quad
Bernard Ghanem$^{1}$
\and
$^{1}$ KAUST
\quad $^{2}$4Paradigm Inc.
\quad $^{3}$Moonshot AI
\quad $^{4}$NVIDIA
}

\begin{document}
\maketitle
\input{sec/0_abstract}    
\input{sec/1_intro}

\input{sec/2_related}
\input{sec/3_method}

\input{sec/4_experiments}

\input{sec/5_conclusion}

{
    \small
    \bibliographystyle{ieeenat_fullname}
    \bibliography{main}
}

\clearpage
\section{Appendix}
\renewcommand\thesection{\Alph{section}}
\renewcommand\thesubsection{\thesection.\arabic{subsection}}
\setcounter{section}{0}
\input{sec/appendix}


\end{document}

%% file: preamble.tex
\usepackage[accsupp]{axessibility} 
\usepackage{graphicx}
\usepackage{amsmath}
\usepackage{amssymb}
\usepackage{booktabs}

\usepackage{pifont} 

\usepackage{graphicx}
\usepackage{booktabs}
\usepackage{multirow}
\usepackage{makecell}
\usepackage{caption}
\usepackage{threeparttable}
\usepackage{algorithm} 
\usepackage{algorithmic} 
\usepackage{listings}
\usepackage{bm}
\usepackage[dvipsnames]{colortbl}
\usepackage[dvipsnames]{xcolor} 
\definecolor{Gray}{gray}{0.85}
\usepackage{hhline}
\usepackage{arydshln}

\def\ie{\textit{i.e.}}

\definecolor{KleinBlue}{rgb}{0.0, 0.129, 0.7}


%% file: sec/0_abstract.tex
\begin{abstract}

As a fundamental task in long-form video understanding, temporal action detection (TAD) aims to capture inherent temporal relations in untrimmed videos and identify candidate actions with precise boundaries. Over the years, various networks, including convolutions, graphs, and transformers, have been explored for effective temporal modeling for TAD. However, these modules typically treat past and future information equally, overlooking the crucial fact that changes in action boundaries are essentially causal events. 
Inspired by this insight, we propose leveraging the temporal causality of actions to enhance TAD representation by restricting the model's access to only past or future context. We introduce CausalTAD, which combines causal attention and causal Mamba to achieve state-of-the-art performance on multiple benchmarks. Notably, with CausalTAD, we ranked 1st in the Action Recognition, Action Detection, and Audio-Based Interaction Detection tracks at the EPIC-Kitchens Challenge 2024, as well as 1st in the Moment Queries track at the Ego4D Challenge 2024. Our code is available at \url{https://github.com/sming256/OpenTAD}.

\end{abstract}

%% file: sec/1_intro.tex
\section{Introduction}
\label{sec:intro}

Temporal Action Detection (TAD) is a fundamental task in long-form video understanding, aiming to localize candidate actions in untrimmed videos and predict their start time, end time, and category~\cite{caba2015activitynet,yang2023basictad, xu2020g,ramazanova2023owl,zhao2021video}. TAD applications span both first-person~\cite{jiang2014thumos,caba2015activitynet} and third-person perspectives~\cite{grauman2022ego4d, damen2022rescaling}, contributing to various domains such as temporal action segmentation~\cite{li2021temporal}, highlight detection~\cite{yao2016highlight}, and video language grounding~\cite{Soldan_2021_ICCV,soldan2021mad}.

\begin{figure}[t]
\centering
\includegraphics[width=1.0\linewidth]{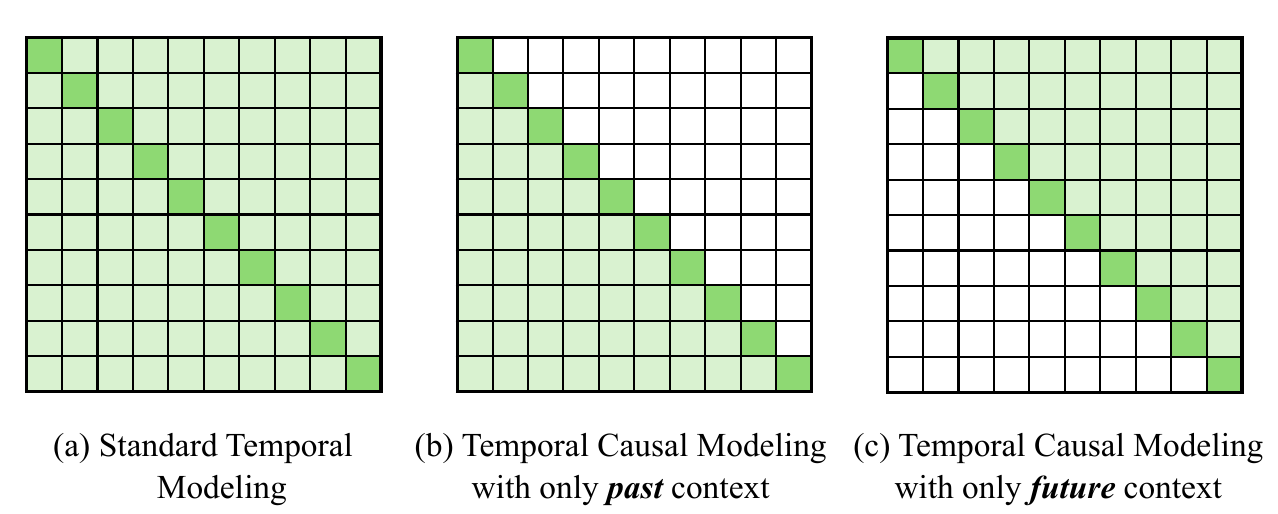}
\caption{(a) Standard temporal modeling treats the past and future context equally, such as convolutions, graphs, and self-attentions, overlooking the fact that changes in action boundaries are essentially causal events. (b) and (c) mitigate this issue by restricting the model's access to only past or future context independently.}
\label{fig:intro}
\end{figure}

Despite its significance, TAD remains challenging due to the complex nature of long-form videos, which contain intricate temporal dependencies and variable-length action instances. Various architectures have been applied for effective temporal modeling, including convolutional networks~\cite{lin2018bsn,lin2019bmn}, graph networks~\cite{xu2020g, zhao2021video}, and recurrent networks~\cite{liu2023etad}. However, these models often struggle to capture long-range context. Recently, transformer-based methods~\cite{zhang2022actionformer} have marked a significant milestone in TAD research, demonstrating the capability to model long-range temporal dependencies.

Nevertheless, these models typically employ non-causal modeling, treating both past and future temporal contexts symmetrically and neglecting the inherent causality in action transitions (Figure~\ref{fig:intro}). In real-world scenarios, action boundaries are usually influenced by causal events. For instance, the end of a running action typically depends on past occurrences, such as reaching a finish line or encountering an obstacle. This suggests a need for models that can understand and leverage temporal causality.
Recent work has utilized causal sequence modeling in the VideoMambaSuite~\cite{chen2024video}, leveraging the Mamba scanning mechanism~\cite{gu2023mamba}. Building on this, we extend the causality-aware design to self-attention and find it effective for various modules. By limiting the valid temporal context to only past or future information, thereby mimicking the natural progression of real-world events, the model can identify action transitions more accurately and improve overall detection performance.

Furthermore, by combining causal attention and causal Mamba in a novel hybrid causal block, we propose the CausalTAD for advanced temporal action detection. The synergy between causal attention and causal Mamba allows for more precise and contextually aware action detection, setting new benchmarks in the field.

Our experiments on multiple datasets demonstrate the superiority of CausalTAD over existing methods and highlight the importance of incorporating temporal causality in TAD. Specifically, we achieve 34.99\% average mAP on the test set of the Ego4D Moment Queries task~\cite{grauman2022ego4d} and 31.97\% average mAP on the test set of the EPIC-Kitchens 100 action detection task~\cite{damen2022rescaling}, ranking first in these tracks at the EgoVis Challenge 2024. We hope our study can shed light on future developments in understanding the temporal dynamics in videos.

We summarize our contributions as follows:

\begin{enumerate}
\item We highlight the importance of incorporating temporal causality in TAD and propose a novel hybrid causal block that leverages causal self-attention and causal Mamba to enhance temporal relationship modeling.
\item Our method achieves state-of-the-art performance across four TAD datasets. Remarkably, with our proposed CausalTAD and stronger video features, we ranked first in the Action Recognition, Action Detection, and Audio-Based Interaction Detection tracks at the EPIC-Kitchens Challenge 2024, as well as first in the Moment Queries track at the Ego4D Challenge 2024.
\end{enumerate}

%% file: sec/2_related.tex
\section{Related Work}

\noindent \textbf{Temporal Action Detection}, also referred to as temporal action localization, can be broadly classified into three categories: one-stage methods~\cite{yang2020revisiting,yang2023basictad,shao2023action}, two-stage methods~\cite{lin2019bmn,wang2020boundary,xia2022learning,Zhao_2023_ICCV}, and DETR-based methods~\cite{tan2021relaxed,liu2022end,shi2022react}. One-stage methods directly localize actions from multi-scale feature pyramids, while two-stage methods require an additional RoI extraction to refine action proposals. Due to their simplicity and strong performance, recent work has primarily focused on one-stage frameworks, such as ActionFormer~\cite{zhang2022actionformer}, TemporalMaxer~\cite{tang2023temporalmaxer}, TriDet~\cite{shi2023tridet}, and VideoMambaSuite~\cite{chen2024video}. Among these, ActionFormer proposes a local transformer block to aggregate temporal context, establishing a powerful one-stage baseline.

TAD methods can also be categorized as feature-based or end-to-end approaches. End-to-end methods, which jointly optimize the video encoder and action detector, have gained increasing attention in recent works~\cite{liu2023etad}. Notably, AdaTAD~\cite{liu2024end} proposes inserting adapters into the backbone while freezing the remaining parameters, achieving high efficiency and state-of-the-art performance on multiple datasets.

Despite these developments, most previous methods ignore temporal causality in TAD, treating past and future information symmetrically. In our work, we follow the feature-based approach and aim to enhance video representation for TAD by leveraging temporal causality.

%% file: sec/3_method.tex
\section{Methodology}
\label{sec:method}

In this section, we first introduce our TAD framework that encompasses feature extraction and action detection. Then, we detail our proposed hybrid causal block, which combines the Mamba and self-attention mechanisms.

\begin{figure*}[t]
\centering
\includegraphics[width=1.0\linewidth]{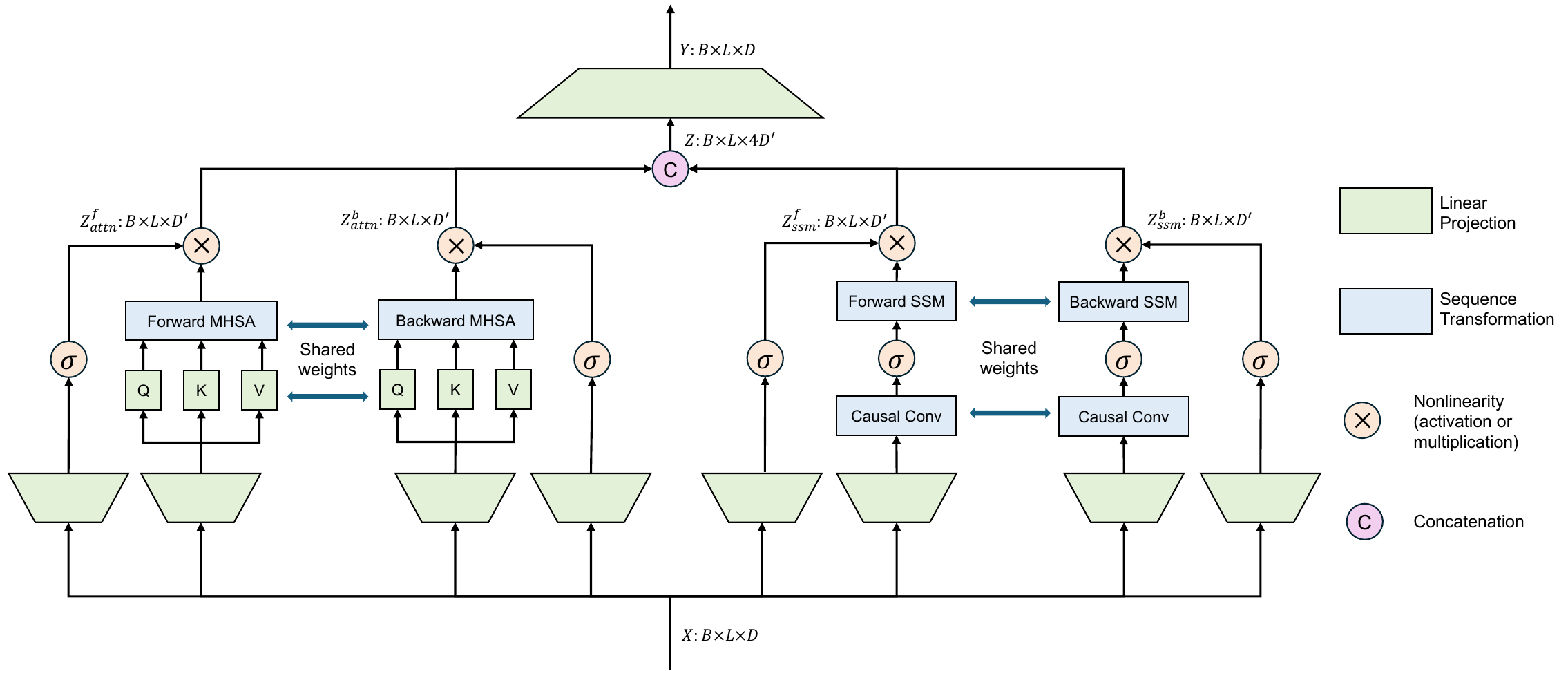}
\caption{\textbf{Hybrid Causal Block.} We combine the Multi-Head Self-Attention (MHSA) and Mamba block (SSM) together, and limit their visible temporal context to only past or future tokens, aiming to capture long-range temporal dependencies and causality. The parameters in the forward and backward MHSA and SSM are shared to reduce the overfitting issue in TAD.}
\label{fig:method}
\end{figure*}

\subsection{One-Stage Detection Framework}

\noindent\textbf{Feature Extraction.} We use a pretrained action recognition model as the video encoder to extract semantically-enriched video features. The untrimmed video is divided into multiple short clips, and we use a sliding window approach to extract features for each clip independently. Each video clip may overlap with others depending on the sliding window stride. Spatial-temporal average pooling is applied after the video backbone to obtain dense video features for each clip.

\vspace{4pt}
\noindent\textbf{Action Detection.} Our detection model is based on ActionFormer, a simple yet effective one-stage anchor-free method~\cite{zhang2022actionformer}. We replace the transformer block in ActionFormer with the following introduced hybrid causal block. To establish a stronger detection baseline on the Ego4D-MQ and EPIC-Kitchens datasets, we primarily optimize four hyperparameters in the detector: the number of feature pyramid levels, the regression loss weight, the probability of input channel dropout, and the number of training epochs.

\subsection{Temporal Causal Modeling}

In ActionFormer, a local transformer is employed within multi-scale feature pyramids to aggregate temporal context and capture temporal dependencies. Recently, VideoMambaSuite~\cite{chen2024video} proposed using Mamba~\cite{gu2023mamba} for video understanding tasks. In temporal action detection, it replaces the local transformer in ActionFormer with a decomposed bidirectional Mamba block (DBM), which captures the temporal context in both forward and backward scanning directions.

The Mamba design focuses on \textit{causal modeling}, emphasizing only previous tokens. In contrast, transformers perform self-attention across the entire sequence. Similarly, convolutions and graphs, common in previous TAD works, treat past and future temporal contexts symmetrically, neglecting the inherent causality in action transitions. In real-world scenarios, action boundaries are typically influenced by causal events. Leveraging temporal causality can enhance the discrimination between background motions and foreground activities, which is crucial for video representation in TAD due to the sensitivity of action boundaries.

Based on this intuition, we propose the \textbf{Hybrid Causal Block}, which combines causal attention and causal Mamba to enhance temporal action detection, as shown in Figure~\ref{fig:method}. These components work synergistically to capture the temporal dependencies and causality of action transitions.

\vspace{12pt}
\noindent \textbf{Causal Mamba Block.}
Mamba~\cite{gu2023mamba} is a representative Structured State-Space Model (SSM)~\cite{gu2021efficiently, smith2022simplified}, inspired by the continuous system that maps input data through an implicit hidden state. It proposes a selection mechanism using a parallel scan and data-dependent SSM layer to efficiently capture long-range dependencies with linear complexity.

However, Mamba was originally designed for 1D sequences under causal relations in language tasks, lacking the capability to observe future information common in vision tasks. To address this, VideoMambaSuite~\cite{chen2024video} proposes the decomposed bidirectional Mamba, which separates the input projector and shares the SSM parameters in both past and future scanning directions, achieving stronger performance than traditional ViM~\cite{zhu2024vision}. We follow this architecture in designing the causal Mamba block, as shown on the right side of Figure~\ref{fig:method}.

\vspace{4pt}
\noindent \textbf{Causal Attention Block.} Although the Mamba block effectively captures temporal context, it models long-range temporal relations implicitly through the SSM layer. To explicitly capture global long-range temporal dependency and temporal causality, we propose the causal attention block, shown on the left side of Figure~\ref{fig:method}. This block shares a similar architectural design to the causal Mamba block, including gated projection and bidirectional MHSA.

The bidirectional MHSA follows the causal modeling in VideoMambaSuite, restricting the context of attention to only past or future information. To implement causal attention using only past context (denoted as backward MHSA), the attention scores are computed as follows:
\[
\text{Attention}(Q, K, V) = \text{softmax}\left(\frac{QK^T}{\sqrt{d_k}} + M\right) V
\]
where \( Q, K, V \) are the query, key, and value matrices, \( d_k \) is the dimension of the key vectors, and \( M \) is the causal mask, which is defined as:

\[
M_{ij} = 
\begin{cases} 
0 & \text{if } i \leq j \\
-\infty & \text{otherwise}
\end{cases}
\]

This ensures that each position in the sequence can only attend to its past and present positions, effectively modeling temporal causality. For causal attention using only future context, we modify $M_{ij}$ to 0 when $i \geq j$ to achieve forward MHSA.

Following~\cite{chen2024video}, input projectors are learned separately in each direction, while the parameters in the forward and backward MHSA, \ie, the weights to compute  $Q$,  $K$, and $V$, are shared to improve the overall detection performance. This can reduce the overfitting issue in TAD, which could be beneficial if the dataset is on a small scale.

\vspace{4pt}
\noindent \textbf{Hybrid Causal Block.} The hybrid causal block integrates the causal Mamba block and the causal attention block to leverage their complementary strengths, allowing for more precise and contextually aware action detection. We compute the causal Mamba block and causal attention block separately, concatenate their outputs, and use a linear layer to reduce the channel dimension to match the input, as shown in Figure~\ref{fig:method}.

\begin{table*}[t]
\centering
\caption{\textbf{Results on ActivityNet-1.3 and THUMOS14}, measured by mAP (\%) at different tIoU thresholds. On ActivityNet-1.3, our prediction is combined with CUHK~\cite{zhao2017cuhk} classification results.}
\small
\setlength{\tabcolsep}{3pt}
\begin{tabular}{lcccc>{\columncolor[gray]{0.9}}cccccccc>{\columncolor[gray]{0.9}}c}
\toprule
\multirow{2}{*}{\textbf{Method}}    & \multicolumn{5}{c}{\textbf{ActivityNet-1.3 }}                                               &      & \multicolumn{7}{c}{\textbf{THUMOS14}}                                        \\
\cline{2-6} 
\cline{8-14} 
                &      Feature               & 0.5 & 0.75 & {0.95} & {Avg.} & & {Feature} &{0.3} & {0.4} & {0.5} & {0.6} & {0.7} & {Avg.} \\
\hline
BMN~\cite{lin2019bmn}& TSN & {50.07} & {34.78} & {8.29} & {33.85} && TSN  &{56.0} & 47.4 & 38.8 & 29.7 & 20.5 & 38.5 \\
{G-TAD}~\cite{xu2020g}  & TSN &{50.36} & {34.60} & {9.02} & {34.09} & & TSN &{54.5} & {47.6} & {40.2} & {30.8} & {23.4} & 39.3  \\
TSI~\cite{liu2020tsi} &TSN & 51.18 & 35.02 & 6.59 & 34.15 &&TSN & 61.0 & 52.1 & 42.6 &  33.2 & 22.4 & 42.3 \\
{VSGN~\cite{zhao2021video}} &TSN &{52.38}  & {36.01}  & 8.37 &{35.07} & & I3D & {66.7} &  {60.4} & {52.4} & {41.0} & {30.4} & 50.2 \\
ActionFormer~\cite{zhang2022actionformer} &TSP &53.50 & 36.20 &8.20 & 35.60  &&I3D &82.1 & 77.8 & 71.0 &59.4 &43.9 & 66.8\\
ASL~\cite{shao2023action}   &TSP & 54.10 & 67.40 & 8.00 & 36.20  && I3D 
 & 83.1 & 79.0 & 71.7 & 59.7 & 45.8 & 67.9  \\
TriDet~\cite{shi2023tridet}   &TSP & 54.70 & 38.00 & 8.40 & 36.80  && I3D & 83.6 & 80.1 & 72.9 & 62.4 & 47.4 & 69.3 \\
\textbf{CausalTAD} & TSP & \textbf{55.62} & \textbf{38.51} & \textbf{9.40}  & \textbf{37.46} & & I3D & \textbf{84.4} & \textbf{80.7} & \textbf{73.5} & \textbf{62.7} & \textbf{47.4} & \textbf{69.7} \\
\bottomrule
\end{tabular}
\label{tab:sota_anet_thumos}
\end{table*}

%% file: sec/4_experiments.tex
\section{Experiments}

\subsection{Datasets and Metrics}

We evaluate our method using four datasets: ActivityNet-1.3~\cite{caba2015activitynet}, THUMOS14~\cite{jiang2014thumos}, EPIC-Kitchens 100~\cite{damen2018scaling}, and Ego4D Moment Queries (Ego4D-MQ)~\cite{grauman2022ego4d}. ActivityNet-1.3 and THUMOS14 consist of web-collected, third-person untrimmed videos, while EPIC-Kitchens and Ego4D-MQ comprise egocentric videos. We follow the standard train/val/test splits for each dataset and report performance accordingly. Following common practice, we use mean Average Precision (mAP) at various IoU thresholds and average mAP as evaluation metrics.

\subsection{Implementation Details}

Our method is implemented using the OpenTAD~\cite{2024opentad} codebase on a single A100 GPU. We employ mixed-precision training and flash attention to accelerate the training process. To achieve stronger performance, we optimize hyperparameters of the detection head, including the number of feature pyramid levels, regression loss weight, probability of input channel dropout, and number of training epochs. Additional experimental configurations can be found in our released code.

For the ActivityNet-1.3 and THUMOS14 datasets, we use the commonly employed TSP~\cite{alwassel2020tsp} feature and the two-stream I3D feature~\cite{carreira2017quo}, respectively. For the Ego4D-MQ dataset, we utilize the InternVideo1~\cite{chen2022internvideo} feature, which is based on the VideoMAE-L and fine-tuned on the Ego4D dataset. On EPIC-Kitchens, we adopt the SlowFast~\cite{slowfast_iccv19} feature, which is also fine-tuned on the EPIC dataset.

\subsection{Comparison with SoTA Methods}

Table~\ref{tab:sota_anet_thumos} presents a comparison of our approach with other state-of-the-art (SoTA) methods on the ActivityNet-1.3 and THUMOS14 datasets. Using the same video features, our method achieves 37.46\% average mAP on ActivityNet and 69.7\% average mAP on the THUMOS14 dataset, surpassing the previous SoTA method TriDet~\cite{shi2023tridet}.

We also present our results on the validation set of the Ego4D Moment Queries v2.0 dataset in Table~\ref{table:ego4d_val}. Notably, we improve the average mAP from the previous best of 23.29\% to 32.19\% on this dataset. Detailed ablation studies on this dataset can be found in Section~\ref{sec:ablation}.

\begin{table}[t]
\caption{\textbf{Results on the \textit{validation} set of Ego4D Moment Queries v2.0.} InternVideo1~\cite{chen2022internvideo} is used for feature extraction,  which is pretrained and fine-tuned on Ego4D Moment Queries.}
\label{table:ego4d_val}
\centering
\small
\setlength{\tabcolsep}{3pt}
\begin{tabular}{lccccc>{\columncolor[gray]{0.9}}c}
\toprule
 \textbf{Method}   & \textbf{0.1} & \textbf{0.2} & \textbf{0.3}  & \textbf{0.4} & \textbf{0.5} &\textbf{Avg.}\\
\midrule
VSGN~\cite{zhao2021video}            & -& -& -& - & -  & 19.35  \\
ActionFormer~\cite{zhang2022actionformer}    & -& -& - & -& - & 23.29 \\
\textbf{CausalTAD} & \textbf{37.68} & \textbf{35.28} & \textbf{32.23} & \textbf{29.49} & \textbf{26.29} & \textbf{32.19} \\
\bottomrule
\end{tabular}
\end{table}

\begin{table}[t]
\caption{\textbf{Results on the \textit{validation} set of EPIC-Kitchens 100.} For comparison, SlowFast-R50 is used in all methods. }
\label{table:epic}
\small
\setlength{\tabcolsep}{5pt}
\begin{tabular}{lccccc>{\columncolor[gray]{0.9}}c}
\toprule
\textbf{Method}  & \textbf{0.1} & \textbf{0.2} & \textbf{0.3} & \textbf{0.4} & \textbf{0.5} &\textbf{Avg.} \\
\midrule
\textit{Verb Task} \\
\hline
BMN~\cite{lin2019bmn}   & 10.8 & 8.8 & 8.4 & 7.1& 5.6 & 8.4 \\
G-TAD~\cite{xu2020g} &  12.1 & 11.0 & 9.4 & 8.1 & 6.5 & 9.4 \\
ActionFormer~\cite{zhang2022actionformer}  & 26.6 & 25.4 & 24.2 & 22.3 & 19.1 & 23.5 \\
ASL~\cite{shao2023action}  & 27.9 & - & 25.5 & - & 19.8 & 24.6 \\
TriDet~\cite{shi2023tridet}  &28.6 & 27.4  & 26.1 & 24.2  & 20.8  & 25.4 \\
\textbf{CausalTAD} & \textbf{29.6} & \textbf{28.7} & \textbf{27.2} & \textbf{25.2} & \textbf{21.4} & \textbf{26.4} \\
\hline
\hline
\textit{Noun Task} \\
\hline
BMN ~\cite{lin2019bmn}  & 10.3 & 8.3 & 6.2 & 4.5 & 3.4 & 6.5 \\
G-TAD~\cite{xu2020g}  & 11.0 & 10.0 & 8.6 & 7.0 & 5.4 & 8.4 \\
ActionFormer~\cite{zhang2022actionformer}  & 25.2 & 24.1 & 22.7 & 20.5 & 17.0 & 21.9 \\
ASL~\cite{shao2023action}  & 26.0 & - & 23.4 & - & 17.7 & 22.6 \\
TriDet~\cite{shi2023tridet}  & 27.4  & 26.3  & 24.6 & 22.2  & 18.3  & 23.8 \\
\textbf{CausalTAD} & \textbf{28.1} & \textbf{26.8} & \textbf{25.2} & \textbf{22.6} & \textbf{18.7} & \textbf{24.3} \\
\bottomrule
\end{tabular}
\label{tab:sota_epic}
\end{table}

\begin{table*}[t]
\centering
\caption{\textbf{Ablation on the \textit{validation} and \textit{test} set of Ego4D Moment Queries v2.0.} Note that all the results on the test set are trained using the \textit{train+val} data and ensemble 3 different seeds. Both InternVideo1 and InternVideo2 are fine-tuned in the Ego4D-Verb dataset and Ego4D-MQ dataset. More details can be found in the Appendix.}
\small
\begin{tabular}{ll|cc}
\toprule
\textbf{Method} & \textbf{Feature} &  \textbf{Avg. mAP on \textit{Validation}} & \textbf{Avg. mAP on \textit{Test}}  \\ 
\midrule 
ActionFormer~\cite{sui2023nms}            & InternVideo1 & 27.52 & 28.44 \\ 
Improved Baseline       & InternVideo1 & 29.45 & -\\
\midrule
Causal Attention Block      & InternVideo1 & 30.87 & - \\
Causal Mamba Block            & InternVideo1 & 31.36 & 31.56 \\
\textbf{Hybrid Causal Block}     & InternVideo1 & \textbf{32.19} & \textbf{31.92} \\
\midrule
\textbf{Hybrid Causal Block}     & InternVideo2 & 33.05 & 34.12 \\
\textbf{Hybrid Causal Block}     & InternVideo1 + InternVideo2 & \textbf{33.59} & \textbf{34.99} \\
\bottomrule
\end{tabular}
\label{tab:results}
\end{table*}

Our results on the EPIC-Kitchens dataset are summarized in Table~\ref{tab:sota_epic}. For a fair comparison, SlowFast-R50~\cite{slowfast_iccv19} is used as the feature extractor in all methods. Our approach achieves new state-of-the-art performance on both verb and noun detection tasks, affirming the efficacy of temporal causal modeling in TAD tasks.

\subsection{Ablation Study}
\label{sec:ablation}

We conducted an ablation study on the Ego4D-MQ dataset, as presented in Table~\ref{tab:results}. To achieve stronger performance, all results on the test set were trained using \textit{train+val} data. Additionally, we independently ran experiments three times using the same configuration but with different seeds, and further averaged the classification logits and regression predictions as the ensemble on the test set.

Initially, by optimizing the detection model's hyperparameters using the InternVideo1 features, we observed an increase in mAP from 27.52\% to 29.45\% on the validation set. Substituting the local transformer in ActionFormer with our proposed causal attention block further increased the mAP to 30.87\%, confirming the efficacy of bidirectional temporal causal modeling in TAD tasks. Moreover, the causal Mamba block outperformed the causal attention block, highlighting the advantages of SSM in scenarios with limited video data. Ultimately, our novel hybrid causal block achieved an mAP of 32.19\% on the validation set and 31.92\% on the test set.

Switching to the InternVideo2 feature further boosted performance to 34.12\% mAP on the test set, emphasizing the importance of scaled-up data and models during self-supervised pretraining. In the end, by combining the InternVideo1 and InternVideo2 features, we achieved an average mAP of 34.99\% and a Recall@1x of 52.83\% at tIoU=0.5 on the test set, ranking first in the Ego4D Moment Query Challenge 2024.

We also investigated the impact of context length on egocentric video understanding. Reducing the context length of self-attention and the Mamba block (\ie, using sliding window self-attention or sliding window Mamba) consistently decreased detection performance on the Ego4D dataset. Optimal results were achieved only when utilizing full-sequence length tokens, underscoring the unique importance of long-range temporal modeling in this dataset.

%% file: sec/5_conclusion.tex
\section{Detailed Solution to EgoVis Workshop 2024}

Please refer to the appendix for more details about our solution to the EgoVis Workshop 2024. This includes our approaches for Action Recognition, Action Detection, and Audio-Based Interaction Detection tasks in the EPIC-Kitchens Challenge, as well as the Moment Queries task in the Ego4D Challenge.

\section{Conclusion}

Our key contribution in this report is the proposed hybrid causal block for effective temporal modeling in TAD tasks. The ablation studies and experiments have demonstrated its effectiveness across four datasets. We hope that our approaches and findings can provide valuable insights into the field of long-form egocentric video understanding.

One limitation of our work is that we rely solely on offline features rather than end-to-end training. Due to the large backbone model and the immense number of video frames, achieving efficient end-to-end training remains a significant challenge that warrants further investigation.

\section{Acknowledgments}
This work was supported by the KAUST Office of Sponsored Research through the GenAI Center of Excellence funding, as well as, the SDAIA-KAUST Center of Excellence in Data Science and Artificial Intelligence (SDAIA-KAUST AI).

%% file: sec/appendix.tex
In this appendix, we provide more details about our solution to the EgoVis Workshop 2024, including the Moment Queries task in the Ego4D Challenge in Section~\ref{sec:appendix_ego4d}, and the Action Recognition, Action Detection, and Audio-Based Interaction Detection tasks in the EPIC-Kitchens Challenge in Section~\ref{sec:appendix_epic}.

\section{Ego4D Moment Queries Challenge 2024}
\label{sec:appendix_ego4d}

As shown in Table~\ref{tab:results}, InternVideo1 and InternVideo2 are used to encode powerful video representations in this task. Below, we explain the detailed process to obtain these backbones and details about the feature extraction.

\subsection{Backbone Fine-Tuning} 

Both InternVideo1 and InternVideo2 utilize self-supervised pretraining followed by supervised fine-tuning on the Kinetics-700 dataset~\cite{carreira2019short}. Given the significant domain gap between the third-person perspective videos used in pretraining and the egocentric videos in the MQ dataset, we fine-tune these backbones on the Ego4D dataset. Specifically, we follow a two-stage fine-tuning approach~\cite{chen2022internvideo}, including Ego4D-Verb fine-tuning and MQ fine-tuning, to fine-tune the InternVideo2$_{s1}$-1B backbone.

\vspace{4pt}
\noindent {\textbf{Ego4D-Verb Fine-Tuning.}} 
We use clip-level annotations from EgoVLP~\cite{lin2022egocentric} as the training set and perform action recognition on it, following~\cite{chen2022internvideo}. Although this annotation includes both noun and verb subsets, we find that training on the verb subset better aids the downstream MQ tasks, as MQ contains more action labels. The training set for the verb task comprises 3.18M clips.

For fine-tuning InternVideo2 on Ego4D-Verb, we use AdamW as the optimizer with a weight decay of 0.05, an initial learning rate of 1e-4, and the cosine learning rate decay strategy. The batch size is 240, with a warm-up epoch of 1 and a total training epoch of 8. For each video, we use sparse sampling to sample 16 frames and resize the short side to 224 pixels.

\vspace{4pt}
\noindent \textbf{MQ Fine-Tuning.} 
After Ego4D-Verb fine-tuning, we treat actions within untrimmed videos in the MQ dataset as individual clips, resulting in an action recognition dataset comprising 13,559 videos across 110 classes. We exclude background clips as they decreased downstream detection performance.

For MQ fine-tuning on InternVideo2, we use a learning rate of 2e-4 and a batch size of 128. The warm-up epoch is 5, and the total training epoch is 10. The same sparse sampling technique is utilized to extract 16 frames per clip.

\subsection{Feature Extraction}
Using the MQ-finetuned model, we extract features from the last layer of the backbone using a sliding window approach for untrimmed video. Each snippet spans 16 frames, with a stride of 8 frames between consecutive snippets.

\subsection{Temporal Action Detection}

Our detailed ablation and final results on the Ego4D-MQ dataset are presented in Table~\ref{tab:results}. We use the \textit{train+val} data as training samples and ensemble three different seeds to submit our results to the test server. The sigma threshold in NMS during post-processing is set to 2.0 for the detection task and 0.7 for the retrieval task. In the end, we achieved an average mAP of 34.99\% and a Recall@1x of 52.83\% at tIoU=0.5 on the test set, ranking first in the Ego4D Moment Query Challenge 2024.

\section{EPIC-Kitchens Challenge 2024}
\label{sec:appendix_epic}

\subsection{Action Recognition Task}
\label{sec:ar}

A robust action recognition model is crucial for downstream tasks, such as action detection. Therefore, we have devoted considerable effort to developing a strong recognition model for the EPIC-Kitchens dataset. Our solution builds upon InternVideo2, and we explore ensemble strategies to achieve higher top-1 accuracy.

\subsubsection{Fine-Tuning InternVideo2}

Following the previous experience, we fine-tune the InternVideo2$_{\text{s1}}$-1B model on the verb and noun subsets individually.
Additionally, to enhance the prediction diversity during the ensemble process, we fine-tune another InternVideo2 model to predict action labels directly. For simplicity, these models will be referred to as InternVideo2$_{\text{s1}}$-1B$_{\text{sep}}$ and InternVideo2$_{\text{s1}}$-1B$_{\text{act}}$, respectively.

We select the training hyperparameters based on performance on the validation set and then perform fine-tuning directly on the \textit{train+val} data using these optimized hyperparameters. For each video, we use sparse sampling to extract 16 frames, with the short edge of sampled videos resized to 288 pixels. The model is fine-tuned for 10 epochs with a batch size of 128. We use AdamW as the optimizer with a weight decay of 0.05 and an initial learning rate of 1e-4, employing a cosine learning rate decay strategy. The model is warmed up for 3 epochs. During inference, we sample 4 segments and 3 crops for each video.

\subsubsection{Model Ensemble}

To further improve accuracy on the action task, we combine InternVideo2's predictions with those of LAVILA~\cite{zhao2023lavila}, which enhances representations by leveraging pretrained large language models. We use a simple yet effective strategy to increase the diversity of model predictions. For models trained to predict action labels directly, \ie, the LAVILA and InternVideo2$_{\text{s1}}$-1B$_{\text{act}}$ model, we directly use the 3806-dimension predictions. As for InternVideo2$_{\text{s1}}$-1B$_{\text{sep}}$, we first select the scores of the possible 3806 actions from the 97$\times$300-dimension action probability matrix and then normalize the scores. Finally, we obtain the prediction results by performing a weighted summation of these predictions from different models.

\subsubsection{Results}

\noindent \textbf{Accuracy on the validation set.} 
We present the verb and noun top-1 accuracy of our models on the validation set in Table~\ref{table_epic_ar_val}. The LAVILA-L model, obtained from the official GitHub repository, achieves 65.4\% top-1 noun accuracy and 73.0\% top-1 verb accuracy. Our fine-tuned InternVideo2$_{\text{s1}}$-1B$_{\text{sep}}$ achieves 70.5\% top-1 noun accuracy and 77.6\% top-1 verb accuracy, surpassing the previous state-of-the-art by a significant margin.

\begin{table}[ht]
\centering
\small
\caption{\textbf{Results of Top-1 accuracy on the validation set of EPIC-Kitchens 100 action recognition task.}}
\label{table_epic_ar_val}
\begin{tabular}{lcc}
\toprule
\textbf{Model}  & \textbf{Noun}   & \textbf{Verb} \\ 
\midrule
LAVILA-L                                       & 65.4 & 73.0 \\ 
\textbf{InternVideo2$_{\text{s1}}$-1B$_{\text{sep}}$} & \textbf{70.5} & \textbf{77.6} \\ 
\bottomrule
\end{tabular}
\end{table}

\noindent \textbf{Accuracy on the test set.} Table~\ref{table_ar_test} summarizes the accuracy of our models on the EPIC-Kitchens test set. Our fine-tuned InternVideo2$_{\text{s1}}$-1B$_{\text{sep}}$ achieves a top-1 noun accuracy of 68.4\%, top-1 verb accuracy of 73.5\%, and top-1 action accuracy of 55.3\%. Next, we explore different ensemble strategies to enhance the performance on the action task.

In Table~\ref{table_ar_test}, Ensemble 1 refers to the combination of InternVideo2$_{\text{s1}}$-1B$_{\text{sep}}$, InternVideo2$_{\text{s1}}$-1B$_{\text{act}}$, and LAVILA-L models, each contributing equally with a weight ratio of 1:1:1. Additionally, we observed that ten epochs were insufficient for the InternVideo2$_{\text{s1}}$-1B$_{\text{act}}$ model, prompting us to fine-tune another model of the same type for 20 epochs. Ensembles 2 and 3 incorporate these four action recognition models, differing only in their weighting ratios. Ensemble 2 utilizes an equal weighting of 1:1:1:1. For Ensemble 3, we adjust the weights by reducing that of InternVideo2$_{\text{s1}}$-1B$_{\text{act}}$ and increasing the weight of InternVideo2$_{\text{s1}}$-1B$_{\text{sep}}$, resulting in a final ratio of 0.3:0.225:0.225:0.25. Our ensemble policy ultimately achieves a top-1 action accuracy of 57.9\% on the test set, securing first place in the competition.

\begin{table}[ht]
\centering
\small
\caption{\textbf{Results of Top-1 accuracy on the test set of EPIC-Kitchens 100 action recognition task.}}
\vspace{-4pt}
\label{table_ar_test}
\begin{tabular}{lccc}
\toprule
\textbf{Model}  & \textbf{Action}  & \textbf{Noun}   & \textbf{Verb} \\ 
\midrule                               
InternVideo2$_{\text{s1}}$-1B$_{\text{sep}}$ & 55.3   & 68.4 & 73.5 \\ 
\midrule                               
Ensemble 1                                   & 57.6   & -    &  -    \\ 
Ensemble 2                                   & 57.7   & -     &  -    \\
Ensemble 3                                   & \textbf{57.9}   & -    &  -    \\ 
\bottomrule
\end{tabular}
\end{table}

\subsection{Temporal Action Detection}

\subsubsection{Feature Extraction}
We employ the previously fine-tuned InternVideo2$_{s1}$-1B$_{\text{sep}}$ as the video encoding backbone, but trained without validation videos to avoid overfitting. Note that we use a sliding window approach to extract snippet features of nouns and verbs separately, with each snippet spanning 16 frames and a stride of 8 frames between consecutive snippets.

\subsubsection{Detection Model}

We use the same detection model, CausalTAD, as introduced in Section~\ref{sec:method} for temporal action detection on EPIC-Kitchens. We train the noun and verb models separately using their corresponding ground truth. Although these two models can be evaluated on noun and verb tasks separately, further post-processing is needed to generate predictions for the action task under joint evaluation.

To construct the action predictions, we select the top-10 noun classes and top-10 verb classes for each timestamp and calculate their product to establish candidate action probabilities. We then apply Soft-NMS~\cite{softNMS} to eliminate redundant proposals.

Using these action labels, the detection performance for the noun and verb tasks decreases by approximately 3-5\% mAP, as shown in Table~\ref{table_epic_val}. This indicates a potential conflict between these two tasks. For example, the mAP for the noun task decreased from 37.02\% to 34.44\%. 

\begin{table}[h]
\centering
\small
\caption{\textbf{Comparison of separate evaluation and joint evaluation on the validation set of EPIC-Kitchens 100 action detection task.} InternVideo2 feature is used.}
\vspace{-4pt}
\label{table_epic_val}
\begin{tabular}{lccc}
\toprule
\textbf{Model}  & \textbf{Noun}   & \textbf{Verb} & \textbf{Action} \\ 
\midrule
Separate Evaluation                                       & 37.02 & 32.30 & -\\ 
\textbf{Joint Evaluation} & \textbf{34.44} & \textbf{27.67} & \textbf{29.63} \\ 
\bottomrule
\end{tabular}
\vspace{-8pt}
\end{table}

\begin{table*}[t]
\centering
\small
\caption{\textbf{Results on the EPIC-Kitchens 100 action detection task.}  All results on the test set are trained using \textit{train+val} data without ensemble, and then evaluated on the test server. }
\label{table:all epic kitchens ablation}
\begin{tabular}{cccccccccc}
\toprule
    \multirow{2}{*}{\textbf{Split}}  & \multirow{2}{*}{\textbf{Method}}  & \multirow{2}{*}{\textbf{Feature}} & \multirow{2}{*}{\textbf{Task}} & \multicolumn{6}{c}{\textbf{mAP@tIoU}}  \\ 
     \cline{5-10}& & & &\textbf{0.1} & \textbf{0.2} & \textbf{0.3} & \textbf{0.4} & \textbf{0.5} & \textbf{Avg.} \\ 
   \midrule 
   \multirow{3}{*}{\textbf{Val}}  
  & \multirow{3}{*}{CausalTAD} &  \multirow{3}{*}{InternVideo2-1B} 
          & Verb         & 31.05 & 29.96 & 28.02 & 25.75 & 23.60 & \textbf{27.67}\\ 
   & & & Noun            & 39.36 & 37.78 & 35.53 & 32.12 & 27.42 & \textbf{34.44}\\ 
   & & & \textit{Action} & 33.01 & 32.03 & 30.28 & 27.86 & 24.98 & \textbf{29.63} \\
   
   \hline 
  \multirow{3}{*}{\textbf{Test}}  
    & \multirow{3}{*}{CausalTAD} &  \multirow{3}{*}{InternVideo2-1B} 
       & Verb            & 35.79 & 34.10 & 30.48 & 28.02 & 24.73 & \textbf{30.02}\\
   & & & Noun            & 40.66 & 38.62 & 36.31 & 32.54 & 27.98 & \textbf{35.22}\\
   & & & \textit{Action} & 36.09 & 34.69 & 32.67 & 29.91 & 26.50 & \textbf{31.97}\\
\bottomrule
\end{tabular}
\label{tab:tad_final}
\end{table*}

\subsubsection{Results}

The conclusive results for noun, verb, and action tasks on the validation and test subsets are reported in Table~\ref{tab:tad_final}. Ultimately, we achieve an average mAP of 31.97\% on the action task in the test set.

\subsection{Audio-Based Interaction Detection}

\subsubsection{Detection Model}

Audio-based interaction detection aims to localize candidate actions within untrimmed videos, emphasizing audio cues as the primary indicators of target actions. Following the design outlined in Section~\ref{sec:method}, we employ CausalTAD as the detector for enhanced temporal modeling on this task.

Uniquely, rather than relying on InternVideo2 features, we prioritize audio cues, utilizing the Auditory-SlowFast model~\cite{kazakos2021slow} for feature extraction. Auditory-SlowFast is a dual-stream convolutional network designed for audio recognition, processing time-frequency spectra, and pretrained on the EPIC-Kitchens action recognition task.

\subsubsection{Results}

Our implementation is based on the OpenTAD framework~\cite{2024opentad}. We present the results of our audio-based interaction detection in Table~\ref{tab:audio_tad}. Compared to the baseline, which is ActionFormer, we significantly improve the average mAP from 7.35\% to 14.82\% on the test set. This enhancement confirms the effectiveness of our detection model, establishing a solid baseline for audio-based interaction detection.

\begin{table}[h]
\centering
\small
\setlength{\tabcolsep}{2.5pt}
\caption{\textbf{Results on the EPIC-Kitchens 100 audio-based interaction detection task.} All the methods use the Auditory-SlowFast feature for fair comparison. Our results on the test set are trained using \textit{train+val} data without ensemble. }
\begin{tabular}{lccccccc}
\toprule
\textbf{Method}  & \textbf{Subset} & \textbf{0.1} & \textbf{0.2} & \textbf{0.3} & \textbf{0.4} & \textbf{0.5} & \textbf{Avg.}\\
\midrule 
CausalTAD & Val  & 16.85 & 15.64 & 14.60 & 12.99 & 11.22 & 14.26 \\ 
\midrule
ActionFormer & Test  & 9.57 & 8.51 & 7.38 & 6.22 & 5.05 & 7.35 \\
CausalTAD & Test  & \textbf{19.81} & \textbf{17.24} & \textbf{14.82} & \textbf{12.48} & \textbf{9.74} & \textbf{14.82} \\
\bottomrule
\end{tabular}
\label{tab:audio_tad}
\end{table}

Additionally, we ablate the impact of visual features on the audio-based interaction detection task in Table~\ref{tab:audio_tad_ablation}. Although the visual backbone InternVideo2 has already been fine-tuned on the EPIC dataset and achieves great performance on the standard action detection dataset, we find its performance on this audio-based benchmark is not satisfactory, lagging behind the audio-pretrained backbone. This further confirms the importance of the audio modality on this dataset. With stronger audio features and additional visual features, we expect higher detection performance.

\begin{table}[t]
\centering
\small
\setlength{\tabcolsep}{3.2pt}
\caption{\textbf{Ablation of different features on the validation set of EPIC-Kitchens 100 audio-based interaction detection task.}}
\begin{tabular}{lcccccc}
\toprule
\textbf{Feature}  & \textbf{0.1} & \textbf{0.2} & \textbf{0.3} & \textbf{0.4} & \textbf{0.5} & \textbf{Avg.}\\
\midrule 
InternVideo2-Verb  & 13.28 & 11.77 & 9.39 & 7.55 & 5.91 & 9.58 \\ 
InternVideo2-Noun  & 14.80 & 13.06 & 11.41 & 9.31 & 6.69 & 11.05 \\ 
\textbf{Auditory-SlowFast}  & \textbf{16.85} & \textbf{15.64} & \textbf{14.60} & \textbf{12.99} & \textbf{11.22} & \textbf{14.26} \\
\bottomrule
\end{tabular}
\label{tab:audio_tad_ablation}
\end{table}